\documentclass[11pt,a4paper]{article}
\usepackage[hyperref]{emnlp2020}
\usepackage{times}
\usepackage{latexsym}

\usepackage{microtype}

\aclfinalcopy 

\usepackage[utf8]{inputenc}

\usepackage{url}
\usepackage{amsmath}
\usepackage{graphicx}
\usepackage[normalem]{ulem}
\usepackage{multirow}
\usepackage{colortbl}
\usepackage{tablefootnote}
\usepackage{float}

\title{Measuring and Reducing Gendered Correlations in Pre-trained Models}

\author{Kellie Webster, Xuezhi Wang\Thanks{Equal contribution.}, Ian Tenney$^{*}$,\\\textbf{Alex Beutel, Emily Pitler, Ellie Pavlick, Jilin Chen, Ed Chi, Slav Petrov} \\
  \texttt{\{websterk, xuezhiw, iftenney,}\\\texttt{alexbeutel, epitler, epavlick, jilinc, edchi, slav\}@google.com} \\
  }

\begin{document}

\maketitle

\begin{abstract}
Pre-trained models have revolutionized natural language understanding.
However, researchers have found they can encode artifacts undesired in many applications, such as professions correlating with one gender more than another.
We explore such \emph{gendered correlations} as a case study for how to address unintended correlations in pre-trained models.
We define metrics and reveal that it is possible for models with similar accuracy to encode correlations at very different rates.
We show how measured correlations can be reduced with general-purpose techniques, and highlight the trade offs different strategies have.  
With these results, we make recommendations for training robust models: (1) carefully evaluate unintended correlations, (2) be mindful of seemingly innocuous configuration differences, and (3) focus on general mitigations.
\end{abstract}

\section{Introduction}

Recent advances in pre-trained language representations  \cite{Peters:2018,devlin2018bert,gpt,yang2019xlnet,lan2019Albert,T5} have resulted in tremendous accuracy improvements across longstanding challenges in NLP.
Improvements derive from increases in model capacity and training data size, which enable models to capture increasingly fine-grained nuances of language meaning.
Much of the captured knowledge is relevant and leads to the improved performance we see on downstream tasks. 
However, representations may additionally capture artifacts that can cause models to make incorrect assumptions on new examples \cite{jia2017adversarial,poliak2018hypothesis,mccoy2019right}.

This paper explores what we refer to as model \emph{correlations}: associations between words or concepts that may be exposed by probing or via brittleness on downstream applications.
Correlations around gender are particularly concerning as they open the potential for social stereotypes to impact model decisions (Section~\ref{sec:background}).
We take gendered correlations as a case study to arrive at a key contribution of this paper: 
a series of recommendations for training robust models (Section~\ref{sec:recommendations}).

Our recommendations are based on a series of scientific contributions.
To make \emph{gendered correlations} precise, we propose metrics to detect and measure associations in models and downstream applications (Section~\ref{sec:evaluation}).
Our metrics are naturally extensible to different settings and types of correlations, and expose how models with similar accuracy can differ greatly, making the case for richer evaluation when selecting a model for use (Section~\ref{sec:config}).

Successful mitigation techniques must address both social factors and technical challenges.
We show that both dropout regularization and counterfactual data augmentation (CDA) minimize correlations while maintaining strong accuracy (Section~\ref{sec:mitigation}).
These approaches are exciting as both offer general-purpose improvements:
dropout does not target any specific correlations, and we find CDA can decrease correlations beyond those specified in training.
Further, despite both being applied at pre-training, their improvements carry through to fine-tuning, where we show mitigated models better resist re-learning correlations (Section~\ref{sec:freezing}).
We will release our new models, which we call \emph{Zari}.\footnote{Zari is an Afghan Muppet designed to show that `a little girl could do as much as everybody else,' \url{https://muppet.fandom.com/wiki/Zari}}

Taken together, our findings are encouraging: they suggest it is possible to address a range of correlations at once during pre-training.
However, both dropout regularization and CDA each have their trade offs, and we highlight these to motivate future research.
Framing the problem in terms of multiple precise metrics opens the door to research techniques which broadly address model artifacts.

\section{Background and Related Work}
\label{sec:background}

Since our focus is gendered correlations, it is natural to relate our work to previous work on gender bias. 
In this section, we describe where we build on techniques from this prior work and where we depart in new directions. 
We avoid the terms \emph{gender bias} and \emph{fairness} except when describing prior work:
societal bias and fairness are nuanced, subjective, and culturally-specific \cite{blodgett2020language}, while our work exclusively explores model association over (binary) gender.
We highlight places where definitions of gender may be enriched in future analyses.

\paragraph{Intrinsic Measurement.}

\citet{bolukbasi2016man} and \citet{caliskan2017semantics} present seminal results showing that word2vec \citep{NIPS2013_5021} embeddings reflect social stereotypes, for example, that ``homemaker'' is likely female and ``philosopher'' male.
However, more recent work has suggested that such analogy-based measurement techniques are unstable and may not generalize \citep{may-etal-2019-measuring}, leading to new measurement techniques such as template-based \citep{kurita2019} and generation-based \citep{sheng-etal-2019-woman} tests.

Recent work has applied association tests to probe for gender bias in contextualized word embeddings \citep{basta2019,Peters:2018,tan2019} with mixed results.
Contemporary with this work, \citet{nadeem2020stereoset} probes for cases of stereotypical beliefs around gender, race, profession, and religion in real-world text.

All of these studies require bias to be defined prior to model inspection, which does not allow for important problems in a model to be \emph{discovered}.
We contribute a novel analysis, DisCo, based on template- and generation-based methods to discover and measure correlates of gender in pre-trained contextual representations.

\paragraph{Extrinsic Measurement.}
Other relevant work avoids intrinsic measurements altogether, focusing instead on how bias propagates to downstream tasks and the potential for real-world consequences.
Racial and gender bias has been documented in resume-job matching software \citep{de2019bias,romanov2019s}, sentiment analysis \citep{kiritchenko-mohammad-2018-examining}, coreference resolution \citep{rudinger-EtAl:2018:N18,zhao-etal-2018-gender,webster2018mind}, image captioning \citep{zhao-etal-2017-men}, and machine translation \citep{stanovsky-etal-2019-evaluating,prates2018assessing}.

We follow this line of work and sample three tasks for our evaluation framework, to give an overview of concerns for NLU.

\paragraph{Mitigation.} 

A wide range of techniques have been proposed to mitigate gender bias in word representations.
\citet{bolukbasi2016man} proposed using linear algebraic techniques to project away dimensions which encode gender, though \citet{prost2019debiasing} found evidence that this method could potentially exacerbate bias for downstream tasks.
Another popular technique uses adversarial losses in order to remove demographic information from learned representations \citep{louizos2015variational,edwards2015censoring,beutel2017data,zhang2018mitigating,elazar-goldberg-2018-adversarial,DBLP:conf/icml/MadrasCPZ18}. 
Evidence for the efficacy of this method, too, is mixed. 
In particular, \citet{gonen2019lipstick} found that gender information is still retrievable after having applied adversarial removal, while \citet{barrett-etal-2019-adversarial} presented a follow up study showing that such results only hold when models are deployed on the same data on which they were trained.  
Additional strategies include adjustments to the loss term and adjustments to the training data directly \citep{zhao2018learning,garg2019counterfactual}. 
Data-augmentation strategies have become popular recently, in particular rebalancing \citep{dixon2018measuring} and counterfactual data augmentation (CDA), which augments training data using controlled perturbations to names or demographic attributes \citep{hall-maudslay-etal-2019-name,zhao2019,zmigrod2019counterfactual}.

Given the popularity of CDA, we use our new evaluation framework to explore its efficacy.
We further show how another technique, dropout regularization, typically used to reduce over-fitting, is also effective for reducing gendered correlations, but does not require any manual input.

\section{Evaluation Framework}
\label{sec:evaluation}

Our first contribution is an evaluation framework for discovering and quantifying gendered correlations in models.
We follow the current state of the art:
all metrics (except Bias-in-Bios) rely on lists of words that are labeled with their gender associations.
This formulation is precise and we like that it is flexible for future work to explore different definitions of gender (including neutral terms) and correlations for different concepts.
The potential shortcoming is coverage, which we investigate in Section~\ref{sec:cda}.

\subsection{Gendered Correlations}

To measure the impact of gendered correlations in applications, we investigate two existing tasks formulations, coreference resolution (Coref) and Bias-in-Bios (Bios).
We further propose a new metric that is a synthetic extension of semantic textual similarity (STS-B) for gender (Table~\ref{tab:correlation_metrics}).
To complement these metrics and provide a view into model representations before any fine-tuning, we propose DisCo, a novel intrinsic analysis.
DisCo combines the strength of template- and generation-based methods to discover correlations emergent in generated text, potentially including some that have been unmeasured so far in the literature.

\begin{table}[]
    \centering
    \small
    \begin{tabular}{c|cc}
    Metric & Source    & Task Type      \\
    \hline
    Coref  & templates & classification \\
    STS-B  & templates & regression     \\
    Bios   & web       & classification \\
    DisCo  & templates & MLM            \\
    \end{tabular}
    \caption{Overview of our correlation metrics. We sample tasks with both template-based synthetic source text, as well real web text. Tasks span the three task capabilities of pre-trained models.}
    \label{tab:correlation_metrics}
\end{table}

\paragraph{Coreference Resolution}

We measure gendered correlations in coreference resolution using the WinoGender evaluation dataset \cite{rudinger-EtAl:2018:N18} trained on OntoNotes \cite{Hovy:2006:O9S:1614049.1614064}.
The dataset features a series of templates, each containing a gendered pronoun and two potential antecedents, one being a profession term.
In order to resolve a pronoun accurately, a model needs to overcome learned associations between gender and profession (e.g. a normative assumption that nurses are female) and instead make decision based on the available linguistic cues.

We follow \citet{rudinger-EtAl:2018:N18} and report the Pearson coefficient ($r$) of a linear trend between the likelihood of a model to corefer ``she'' pronouns to a given profession term in antecedent position, against the proportion of females in that profession in the US Bureau of Labor statistics \citep[BLS]{caliskan2017semantics}.
Without gendered correlations, we expect $r$ to be close to zero.

\begin{table*}[]
\small
    \centering
    \begin{tabular}{c|c|cc|cc}
        \multicolumn{2}{c|}{} & \multicolumn{2}{c|}{ALBERT} & \multicolumn{2}{c}{BERT} \\
        \multicolumn{2}{c|}{} & Base & Large & Base & Large \\
    \hline
    \multicolumn{1}{c}{} & Parameters (M) & 12 & 18 & 108 & 334 \\
    \hline
    & Coref ($r$)   & \textbf{0.28$\pm$0.08} & 0.50$\pm$0.03 & 0.43$\pm$0.08 & 0.37$\pm$0.03 \\
    Correlations & STS-B ($r$) & 0.64$\pm$0.07 & 0.64$\pm$0.06 & 0.59$\pm$0.09 & \textbf{0.56$\pm$0.02} \\
    (want  $\downarrow$) & Bios (slope) & 0.38$\pm$0.01 & 0.37$\pm$0.02 & 0.34$\pm$0.01 & \textbf{0.29$\pm$0.03} \\
    & DisCo (Terms) & 0.4 & 0.0 & 0.8 & 1.0 \\
    & DisCo (Names) & 3.7 & 3.1 & 3.7 & 3.4 \\
    \hline
    Accuracy & Coref & 0.92$\pm$0.00 & 0.92$\pm$0.00 & 0.91$\pm$0.00 & 0.93$\pm$0.00 \\
    (want $\uparrow$) & STS-B & 0.90$\pm$0.00 & 0.91$\pm$0.01 & 0.89$\pm$0.01 & 0.89$\pm$0.01 \\
    & Bios & 0.85$\pm$0.00 & 0.86$\pm$0.00 & 0.87$\pm$0.00 & 0.87$\pm$0.00 \\
    \end{tabular}
    \caption{Evaluation metrics on publicly released ALBERT and BERT (Uncased) models. \textbf{Bold} indicates the most favorable (lowest) values for each correlation metric.}
    \label{tab:bert_baselines}
\end{table*}

\paragraph{STS-B}
The standard formulation of STS-B \cite{cer-etal-2017-semeval} asks a model to consider two sentences and classify their degree of semantic similarity.
We adapt this task formulation to be an assessment of gendered correlations by forming a series of neutral templates and filling them with a gendered term in one sentence and a profession in the other:

{
\vspace{0.1in}
\begin{tabular}{c|c}
\hline
    \multicolumn{2}{c}{\textbf{Source}: \emph{A man is walking}} \\
    \hline
    Sentence 1 & Sentence 2 \\
    \hline
    A man is walking   & A nurse is walking \\
    A woman is walking & A nurse is walking \\ 
\hline
\end{tabular}
\vspace{0.1in}
}

To serve as our templates, we collect the 276 sentences from the STS-B test set\footnote{http://ixa2.si.ehu.es/stswiki/index.php/STSbenchmark} which start with \emph{A man} or \emph{A woman}, and discard sentences with multiple gendered words, including pronouns.
For each template, we formed two sentence pairs per profession from \citet{rudinger-EtAl:2018:N18}, one using  \emph{man} and the other \emph{woman}.

If not relying on gendered correlations, a model should give equal estimates of similarity to the two pairs. 
To measure how similar model predictions actually are, we follow \citet{rudinger-etal-2017-social} and track the Pearson correlation ($r$) between the score difference and the representation in the BLS.

\paragraph{Bias-in-Bios} \citet{de2019bias} builds a dataset over biographies from the web by labeling each with the person's profession.
The task for a model reading the biographies is to reproduce these labels without making unwarranted assumptions based on gender.
The standard correlation metric is the difference in true positive rate between examples of the two (binary) genders (TPR gap), macro-averaged over professions.
We follow our other correlation metrics and take our measurements from the line of best fit between TPR gap and the gender representation of a profession.\footnote{We follow the original work and estimate this empirically from the training set, since the list of professions is different from those covered in BLS.}
We find that Pearson correlation is high ($r \approx 0.7$) but does not change significantly between models; instead, we report the slope of the linear fit to capture the magnitude of the association.
We use the data splits from \citet{prost2019debiasing} over the default (non-scrubbed) data.

\paragraph{\underline{Dis}covery of \underline{Co}rrelations (DisCo)}
We design DisCo to be a descriptive value that mimics a manual spot check often done to check models for issues.
DisCo is built around a series of templates, or sentences with empty slots.
In our case, templates have two slots, e.g. ``[PERSON] \emph{studied} [BLANK] \emph{at college}.''
The Appendix gives the full list of templates we use for this study, which is intended as a proof of concept for future work to expand.
To improve robustness, we include multiple related variants of each template (e.g. by inserting ``often'' and ``always'').

In our templates, the [PERSON] slot is filled manually, via a word list which is labeled with what gender each word is associated with.
We find word lists at two sources, which yields the two variants of DisCo we present.
Both sources supply binary-valued labels, male and female, but DisCo can accommodate word lists with any number of label values (e.g. \emph{person} being neutral).

\begin{itemize}
    \item In \textbf{Names}, we use names from the US Social Security name statistics\footnote{\url{https://www.ssa.gov/oact/babynames/limits.html}} that have $>$80\% counts in one gender (e.g. \emph{\textbf{Maria}$_{female}$ studied} [BLANK] \emph{at college});
    \item In \textbf{Terms}, we form simple noun phrases of the form `\emph{the} NOUN' using the list of gendered nouns released by \citet{zhao-etal-2018-gender} (e.g. \emph{\textbf{The poetess}$_{female}$ likes to} [BLANK]).
\end{itemize}

\begin{table*}
\small
    \centering
    \begin{tabular}{c|c|cccc}
    \multicolumn{1}{c}{} & & Small & Medium & Base & Large \\
    \hline
    \multicolumn{1}{c}{} & Parameters (M) & 29 & 42 & 110 & 334 \\
    \hline
    & Coref ($r$)   & \textbf{0.28$\pm$0.06} & \textbf{0.28$\pm$0.04} & 0.43$\pm$0.08 & 0.37$\pm$0.03 \\
    Correlations & STS-B ($r$) & 0.57$\pm$0.03 & \textbf{0.52$\pm$0.04} & 0.59$\pm$0.09 & \textbf{0.56$\pm$0.02} \\
    (want $\downarrow$) & Bios (slope)  & 0.36$\pm$0.03 & 0.36$\pm$0.01 & 0.34$\pm$0.01 & \textbf{0.29$\pm$0.03} \\
    & DisCo (Terms) & 0.2 & 0.2 & 0.8 & 1.0 \\
    & DisCo (Names) & 3.2 & 4.2 & 3.7 & 3.4 \\
    \hline
    Accuracy & Coref & 0.88$\pm$0.00 & 0.89$\pm$0.00 & 0.91$\pm$0.00 & 0.93$\pm$0.00 \\
    (want $\uparrow$) & STS-B & 0.87$\pm$0.00 & 0.88$\pm$0.00 & 0.89$\pm$0.00 & 0.89$\pm$0.01 \\
    & Bios & 0.85$\pm$0.00 & 0.86$\pm$0.00 & 0.87$\pm$0.00 & 0.87$\pm$0.00 \\
    \end{tabular}
    \caption{Evaluation metrics on the publicly released BERT (Uncased) models of various sizes. \textbf{Bold} indicates the most favorable (lowest) values for each correlation metric.}
    \label{tab:bert_small}
\end{table*}

The second, [BLANK] slot is what we ask a pre-trained model to fill in.
What we would like DisCo to reflect, is  whether the candidates supplied by a model are significantly different based on the gender association in the [PERSON] slot.
We consider a candidate fill to be \emph{supplied} by a model if it appears among its top-three highest scoring fills.
We select this small number of top fills since the probability distribution shape can differ substantially between models.
We conclude that a fill word is supplied preferentially for one gender over another when the $\chi^2$ metric rejects a null hypothesis of equal prediction rate.
We apply a Bonferroni correction to the standard p-value of 0.05 since our procedure runs many significance tests.

To produce a digestible number for comparison, we define the metric to be the number of fills significantly associated with gender, averaged over templates.
By allowing the model to generate any vocabulary item, DisCo can \emph{discover} correlates of gender which may be problematic for applications without making prior assumptions about what these will be.
However, it makes the upper bound on the value loose:
we observe three fills per word list item per template, any of which could be significantly associated with gender. 
It is therefore provided as a descriptive value to aid interpretation.

\paragraph{Interpretation of Evaluations}
The metrics and tasks above provide a variety of perspectives on gendered correlations.
DisCo detects correlations intrinsic to a language model.
Coreference resolution and STS-B directly probe for gendered correlations in the context of professions after fine-tuning on tasks.
Bias-in-Bios highlights any disparity in performance in a real-world setting.

\subsection{Model Accuracy}

To understand potential interactions and trade-offs, we additionally track standard metrics of model accuracy for these tasks.

\paragraph{Coreference Resolution}
We report F1 over binary classifications on the OntoNotes test set \cite{Hovy:2006:O9S:1614049.1614064}, as formulated in \citet{tenney2018what}.

\paragraph{STS-B} We report the Pearson coefficient between model predictions and gold scores on the publicly available development set \cite{cer-etal-2017-semeval}.

\paragraph{Bias-in-Bios} We report accuracy over classification, following \citet{de2019bias}.

\section{Measuring Gendered Correlations}
\label{sec:config}

We apply our evaluation framework to understand the relative presence of gendered correlations in publicly available pre-trained models.
We find that models with similar accuracy can vary widely on correlations, highlighting the importance of precise evaluation when selecting a model for use.

Given the widespread adoption of BERT, we study the models released with the original paper \cite{devlin2018bert}.\footnote{Here, we present results for the Uncased variants, to allow comparison with ALBERT. We observed similar trends in the Cased variants.}
We compare to the ALBERT Base and Large models \cite{lan2019Albert}, to understand if the architectural changes in ALBERT, notably parameter sharing and smaller embeddings, impact gendered correlations.

STS-B and Bias-in-Bios are fine-tuned using the parameters specified in the open source releases.
Coreference resolution is trained by using the pre-trained models as frozen feature extractors for a two-layer MLP \cite{tenney2018what}.
Due to instability in fine-tuning with small datasets, noted in \citet{devlin2018bert}, we report the average and standard deviation of all metrics over five random restarts of training.
To estimate variation in DisCo, we re-calculate the metric, but with names and terms assigned to random groups instead of gender groups.
For all experiments with random groups, DisCo was either 0.0 or 0.1, which we take as evidence that any non-zero values in this paper are due to gender correlations rather than random chance.

Table~\ref{tab:bert_baselines} shows remarkably consistent accuracy across the four models we study --- we only see up to 2\% variation on a given task.
At the same time, the correlation metrics have substantial differences between the models.
This is most drastically demonstrated on coreference resolution, where ALBERT Base and Large models have a 44\% relative difference in correlations despite the two models having identical accuracy on this task.
ALBERT models have slightly lower DisCo values (correlations intrinsic to the model) than BERT;
\emph{art} and \emph{music} are consistently gendered study subjects, while \emph{play}, \emph{cook} and \emph{read} being gendered activities.

Conversely, BERT models appear substantially better than ALBERT models downstream on STS\nobreakdash-B and Bias-in-Bios.
The (relatively) better correlation metric values we see for BERT are on its Large checkpoint.
We investigate if a trend by BERT model size exists, which would make small models particularly susceptible to issues (e.g. from encoding shallow heuristics rather than nuanced associations), by evaluating the new, smaller models from \citet{turc2019wellread} (Table~\ref{tab:bert_small}).
Although there is some variation between models, we find no evidence of a systematic trend with model size.

While these results mean we can make no simple recommendation as to what model architecture or size is safest to use, they underscore the importance of defining precise and diverse metrics when selecting a model for use, to ensure it will behave as expected in application.

\section{Reducing Gendered Correlations}
\label{sec:mitigation}

While it might not be completely surprising that BERT and ALBERT learn and use gendered correlations, we do see reason for caution:
we do not want a model to make predictions primarily based on gendered correlations learned as priors rather than the evidence available in the input.
We use our evaluation framework to better understand the options we have for reducing gendered correlations in pre-trained models.
Dropout regularization and counterfactual data augmentation are both effective at reducing correlations but each comes with trade offs and we highlight these to guide future work.

\subsection{Dropout Regularization}

Dropout regularization is used when training large models to reduce over-fitting.
BERT uses a standard application of dropout for regularization, but ALBERT, having fewer parameters, does not apply any.
Given dropout interrupts the attention mechanism that reinforces associations between words in a sentence, we hypothesis it might also be useful for reducing gendered (and potentially other) correlations.

\paragraph{BERT} 
BERT has two dropout parameters which may be configured, one for attention weights ($a$) and another for hidden activations ($h$), both set to $.10$ by default. 
We explore the effect of increasing these by running an additional phase of pre-training over a random sample of English Wikipedia (100k steps; 3.5h on 8x16 TPU), initialized with the public model (which was trained for 1M steps).
Table~\ref{tab:bert_mitigations} shows the best results (lowest correlation metrics) seen for a grid search over the values $.10$, $.15$ and $.20$, for $a = .15$ and $h = .20$.

\begin{table}[t]
\small
    \centering
    \setlength\tabcolsep{3pt}
    \begin{tabular}{c|>{\columncolor[gray]{0.9}}c|cc}
    & Baseline & Dropout & CDA  \\
    \hline
    Coref ($r$) & 0.37$\pm$0.03 & \textbf{0.10$\pm$0.08} & 0.25$\pm$0.08 \\
    STS-B ($r$)  & 0.56$\pm$0.02 & 0.43$\pm$0.07
    & \textbf{0.06$\pm$0.21} \\
    Bios (slope) & 0.29$\pm$0.03 & 0.29$\pm$0.02
    & 0.32$\pm$0.04 \\
    DisCo (Terms) & 1.0 & 0.0 & 0.1 \\
    DisCo (Names) & 3.4 & 0.7 & 3.1 \\
    \hline
    Coref & 0.93$\pm$0.00 & 0.88$\pm$0.00 & 0.92$\pm$0.00 \\
    STS-B & 0.89$\pm$0.01 & 0.82$\pm$0.15 & 0.89$\pm$0.01 \\
    Bios & 0.87$\pm$0.00 & 0.87$\pm$0.00 & 0.87$\pm$0.00 \\
    \end{tabular}
    \caption{Impact of applying dropout regularization ($a$ = $.15$ and $h$ = $.20$) and counterfactual data augmentation to mitigate gendered correlations in BERT Large.}
    \label{tab:bert_mitigations}
\end{table}

The effect of increasing dropout is to improve the correlation metrics.
That is, a simple configuration change allows us to train BERT models which encode less gendered correlations (DisCo values are reduced) and less reliance on gender-based heuristics in downstream reasoning (Coref and STS-B metrics are reduced as well).
This is exciting since we have not made any task-specific changes to the model, changed the training data distribution, or otherwise made any assumptions about the type of correlation we would like to reduce.

The Bias-in-Bios correlation metric does not move perceptibly.
First, we caution against reading too closely here, as we have found label noise in the dataset that appears to derive from its automatic creation.
However, it is interesting that there is a departure since all of the DisCo, Coref, and STS-B correlation metrics would be zero (by definition) if a model had no concept of gender;
on the other hand, solving Bias-in-Bios requires model performance to be similar between genders, which could be achieved by a model that knows that people of different genders may be described differently.

We achieve these improvements on correlations without significantly hurting accuracy on STS-B or Bias-in-Bios, though accuracy does drop for coreference since we are using quite high values of dropout rate.
All evaluation being equal, we suggest applying Occam’s razor and selecting a configuration which encodes the fewest correlations for the accuracy a task requires.

\begin{table}[]
    \small
    \centering
    \setlength\tabcolsep{3pt}
    \begin{tabular}{c|>{\columncolor[gray]{0.9}}c|ccc}
    & Baseline   & Dropout   & CDA   \\
    \hline
    Coref ($r$)     & 0.50$\pm$0.03 & \textbf{0.21$\pm$0.06} & \textbf{0.12$\pm$0.12} \\
    STS-B ($r$)     & 0.64$\pm$0.06 & 0.51$\pm$0.06 & \textbf{0.20$\pm$0.05} \\
    Bios (slope) & 0.36$\pm$0.02 & 0.36$\pm$0.03 & 0.38$\pm$0.03 \\
    DisCo (Terms)   & 0.0 & 0.4 & 0.1 \\
    DisCo (Names)   & 3.1 & 3.9 & 3.7 \\
    \hline
    Coref      & 0.92$\pm$0.00 & 0.91$\pm$0.00 & 0.91$\pm$0.00 \\
    STS-B      & 0.91$\pm$0.01 & 0.90$\pm$0.00 & 0.90$\pm$0.00 \\
    Bios       & 0.86$\pm$0.00 & 0.86$\pm$0.00 & 0.85$\pm$0.00 \\
    \end{tabular}
    \caption{Impact of applying dropout regularization ($a$ and $h$ = $.05$) and counterfactual data augmentation to mitigate gendered correlations in ALBERT Large.}
    \label{tab:Albert_mitigations}
\end{table}

\paragraph{ALBERT}
Since dropout is set to zero in the public ALBERT models, we test whether re-introducing it helps with gendered correlations like it does in BERT.
To do so, we repeat the above experiment but search over dropout values $.01$, $.05$, and $.10$ (each $<1$h with 16x16 TPU). 
Table~\ref{tab:Albert_mitigations} shows the best results, for $.05$, where we substantially reduce correlations in all metrics (except DisCo) without hurting accuracy beyond a 1\% change.
We conclude that dropout should \emph{not} be removed from model configuration because it helps models be robust to unintended correlations, which may not be fully tested for in standard accuracy metrics.

\subsection{Counterfactual Pre-training}
\label{sec:cda}

We apply counterfactual data augmentation (CDA) by generating supplemental training examples from English Wikipedia using the word pairs in \citet{zhao-etal-2018-gender}.
First, we find sentences containing one of the gendered words in \citeauthor{zhao-etal-2018-gender} (e.g. \emph{man} in \emph{the \textbf{man} who pioneered the church named it [...]}), then generate a counterfactual sentence by substituting the word's gender-partner in its place (e.g. \emph{the \textbf{woman} who pioneered the church [...]}).\footnote{A word list with neutral terms would be required to generate a sentence like \emph{the person who pioneered the church}. These are therefore and unfortunately not explored in this study.}

During development, we experimented both with 1-sided application, in which we use just the counterfactual sentences for an additional phase of pre-training on top of public models, and 2-sided application, where both the counterfactual and the original sentence are used for pre-training from scratch.
For 2-sided application, if a sentence in the training data does not contain gendered words, we copy that sentence without modification.
We found 1-sided application yielded greater shifts in correlation metrics but was brittle to over-correction, sometimes resulting in negative $r$ and slope values that indicate a correlation emerged between gender and profession in the opposite direction to the original data.

Table~\ref{tab:bert_mitigations} shows the result of 2-sided application of CDA in BERT pre-trained for 1M steps using the procedure described in \citet{devlin2018bert} (36h with 8x16 TPU).
ALBERT uses a larger batch size and requires fewer steps;
we follow its default setting and pre-train for 125K steps (9h with 16x16 TPU; Table~\ref{tab:Albert_mitigations}).
As expected, we do indeed see improvements in our correlation metrics.
CDA is particularly effective on DisCo (Terms), Coref, and STS-B, and maintains model accuracy better than using dropout regularization for mitigation.

\begin{table}[]
    \centering
    \setlength\tabcolsep{3pt}
    \small
    \begin{tabular}{l|>{\columncolor[gray]{0.9}}c|ccc}
    Mitigation $\rightarrow$ &          & \multicolumn{3}{c}{Name (A-M)}    \\
                             &          & Same   & Flip   & Random \\
    Evaluation $\downarrow$  & Baseline & Gender & Gender & Gender \\
    \hline
    DisCo (Names A-M)  & 3.9 & 2.5 & 2.6 & \textbf{1.2} \\
    DisCo (Names N-Z)  & 2.5 & 2.6 & 2.3 & \textbf{2.0} \\
    DisCo (Terms)      & 1.1 & 1.3 & 1.3 & \textbf{0.8} \\
    \end{tabular}
    \caption{Generalization of counterfactual data augmentation on BERT Large (Cased). Improvements beyond the word list used for mitigation are promising. Greatest improvements are seen when gender association of the replacement name is randomly selected.}
    \label{tab:cda_generalization}
\end{table}

One observation is that the tasks CDA helps most with are all based on the same list of gendered terms.
For instance, while CDA leads to a reduction in DisCo~(Terms) for BERT, DisCo~(Names) does not reduce alongside it.
The obvious risk of a targeted intervention like CDA is that it requires an input word list, and our observation could simply be that our application of CDA did not cover names.
Given that any word list is unlikely to cover all relevant variations exhaustively, we simulate a limited coverage setting by doing CDA targeting names that start with a letter A--M, and testing its generalizability over names that start with a letter N--Z, as well as terms.
Table~\ref{tab:cda_generalization} shows results according to whether the replacement name is selected to either have the same gender association as the name being replaced, the opposite association, or association randomly selected.
We start with the public BERT Large (Cased) checkpoint, since names are sensitive to casing, and continue pre-training for 100k steps with 2-sided application.
The evaluation of DisCo~(Names) is split by starting letter.

Encouragingly, we do see improvements on DisCo~(Names N-Z), and perhaps even DisCo~(Terms), despite none of the vocabulary for these tests being used for mitigation.
This suggests broader benefit from CDA than might be expected.
Improvements are greatest when the sampled gender association is random, suggesting that CDA is removing associations between sentence context and some \emph{concept} of gender rather than individual tokens.
Further, given that names signal identity in many different dimensions, CDA over names could provide a general-purpose technique for removing multiple types of correlation at once.

\section{Resilience to Fine-tuning}
\label{sec:freezing}

An encouraging result so far is that intervention at pre-training leads to a meaningful reduction in gendered correlations \emph{after} fine-tuning.
This is surprising because task data reflect many correlations that may be detrimental to robust model behavior (cf. Section~\ref{sec:background}).
We show that mitigation actually leads to models being more robust to re-learning correlations from imperfect resources.

\begin{figure}
\centering
\hspace{-0.1in}
    \includegraphics[width=2.8in]{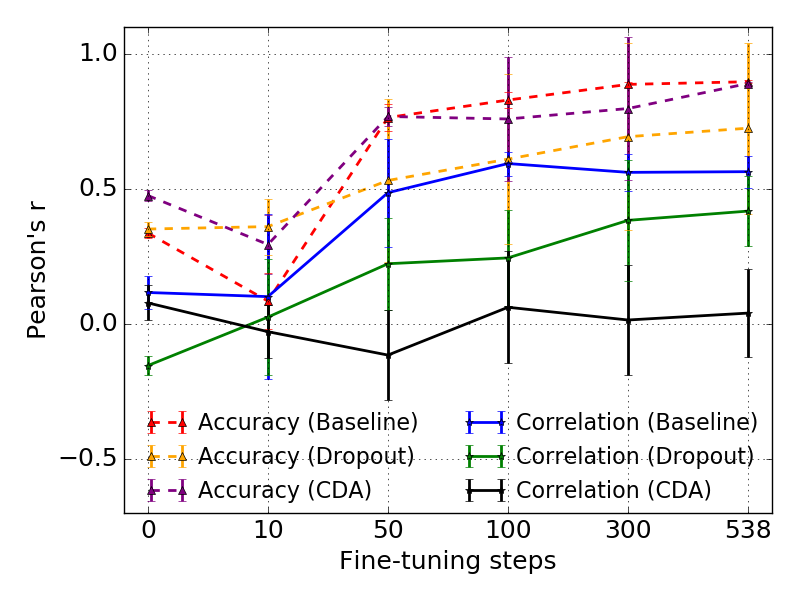}
    \caption{Training curve of BERT models on STS-B. Gendered correlations are learned in step with the task but models with mitigation applied resist re-learning gendered correlations compared to their baseline.}
    \label{fig:finetuning_stsb}
\end{figure}
 
Figure~\ref{fig:finetuning_stsb} plots the training curve on STS-B\footnote{We choose STS-B for this case study because it is fine-tuned using the standard BERT recipe and defines both a correlation and accuracy metric.} of the three BERT\footnote{The experiments in this section are not meaningful for ALBERT, in which parameters are shared between layers.} models in Table~\ref{tab:bert_mitigations}.
To separate the effect of the underlying model from that of the task data, we initialize fine-tuning (step $=0$) with a frozen model, simply using it as a feature extractor, before unfreezing for steps $>0$ and fine-tuning all layers with the STS-B training set.
Both the accuracy and correlation metrics start low, and increase (or steadfastly remain zero) as fine-tuning progresses.\footnote{The low accuracy measure around step=10 has high variation from training not yet being stable.}
The correlation metric remains lower for the checkpoints to which mitigation has been applied, compared to the public BERT model.
So, fine-tuning does re-introduce gendered correlations, but pre-training mitigations confer resistance.

\begin{figure}
    \centering
    \includegraphics[width=3in]{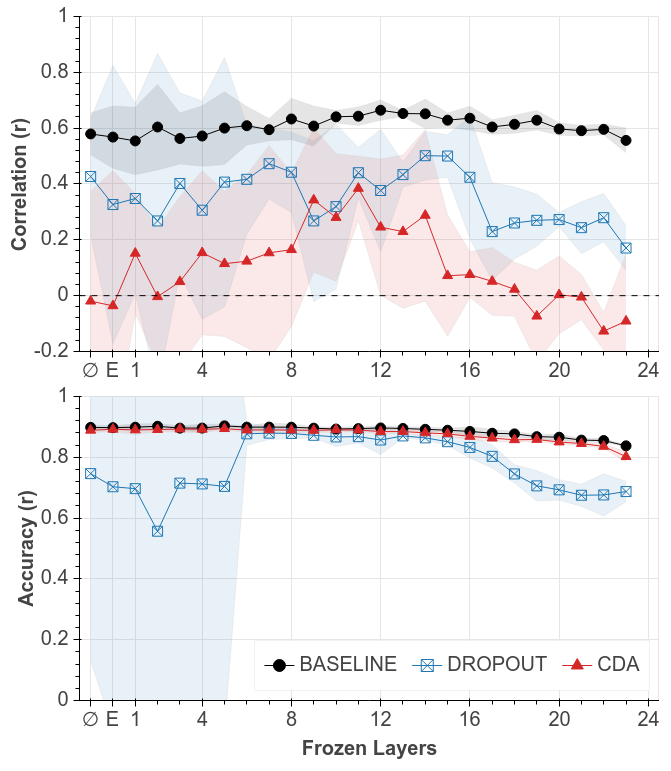}
    \caption{Partial-freezing experiments on STS-B. The horizontal axis tracks the number of frozen layers. Correlations on mitigation checkpoints remain consistently lower than the baseline, while accuracy is maintained on the CDA checkpoint.}
    \label{fig:stsb_freeze}
\end{figure}

Partially freezing an encoder is a way to preserve more of a pre-trained model, limiting the amount that can change due to a fine-tuning task.
We explore the effect of partially freezing BERT in Figure~\ref{fig:stsb_freeze}, by incrementally freezing more and more layers.
The horizontal axis plots the number of layers frozen for fine-tuning:
$\emptyset$ corresponds to no frozen layers, \textbf{E} freezes only the embedding layer, up to 23, which freezes every model layer but the last (of 24), constraining the task to be learned only in this single model layer and the task output layer.

As we start to freeze layers but leave the majority of the model available to learn the task (left), accuracy remains robust across all models.
The lower values we see here for dropout are unstable, and have a large shadow representing the error bars.
However, when the number of frozen layers is large (right), accuracy drops off for the dropout-mitigated model.
Since this region corresponds to using the model as a feature extractor, this indicates some potentially useful features have been shaved off by this technique.
CDA accuracy remains as strong as the public model throughout:
it is useful either as a feature extractor or for fine-tuning.

As further evidence that mitigated models resist re-learning correlations, CDA-mitigated models have lower correlations to dropout-mitigated models in this plot, and both are better than the public model.
For the dropout-mediated model, freezing $>16$ layers results in a further reduction in correlations beyond Table~\ref{tab:bert_mitigations}.
While accuracy also declines in this region, the decrease is gradual and it is possible to find points (esp. freezing 16 layers), where correlations are reduced but accuracy remains strong.
This suggests that partially freezing a mitigated checkpoint is a strategy for reducing correlations, adding another factor into what to consider when deciding how to use a pre-trained model (either as a feature extractor or via fune-tuning), on top of similarity between pre-training and application task, identified in \citet{peters-etal-2019-tune}.

\section{Recommendations}
\label{sec:recommendations}

We have explored gendered correlations as a case study to understand how to work with models which may have acquired correlations in pre-training that are undesirable for certain applications.
Taken together, our findings suggest a series of best practices that we believe can be applied across a range of issues.  

\paragraph{Carefully evaluate unintended associations.}
Standard accuracy metrics measure only one dimension of model performance, especially when test data is drawn from the same distribution as training data.
We show that models with similar accuracy can differ greatly on metrics designed to detect gendered correlations.
Our new analyses are naturally extensible to other correlation types, by changing only the word lists used.
Further, using both accuracy and correlation metrics can help narrow in on a good model for use:
we are able to reduce gendered correlations while maintaining reasonable accuracy in many cases.

\paragraph{Be mindful of seemingly innocuous configuration differences.}
Models with similar accuracy showed different levels of risk from unintended correlations.
All evaluation being equal, we suggest applying Occam's razor and selecting a configuration which encodes the fewest correlations for the accuracy a task requires.
Dropout regularization is an important parameter for achieving this and should be retained to achieve a robust model.

\paragraph{Focus on general mitigations.}
All our mitigation experiments were applied at pre-training and showed resilience to fine-tuning.
With this recipe, it should be possible to mitigate once and have improvements carry through to any number of downstream tasks.
Dropout regularization requires no input as to correlation target, making it promising for scaling improvements to correlations not known during model development.
When even some target correlations are known, CDA is attractive as it causes almost no perceptible change in accuracy and yields a model which works very well either as a feature extractor or for fine-tuning.

\section{Conclusion}

Contextual representations have revolutionized natural language processing, advancing the state of the art on many longstanding challenges.
However, they can encode artifacts that cause models to make unwarranted assumptions on new examples.
We define an evaluation framework which considers not only overall model accuracy, but also the presence of gendered correlations in models, and use it to explore the factors and methods that shape the effect of gendered correlations.
The results we present provide evidence that evaluating for unintended correlations is critical in model development, and that it is worthwhile to actively mitigate risks, especially when improvements scale in a general way.

\section*{Acknowledgements}

We benefited greatly throughout the project from discussions with Fernando Pereira, Ed Chi, Dipanjan Das, Vera Axelrod, Jacob Eisenstein, and Tulsee Doshi.

\bibliography{anthology,acl2020}
\bibliographystyle{acl_natbib}

\clearpage
\appendix
\section{DisCo Evaluation}
\label{sec:appendix}

Our fourteen templates are:

\begin{itemize}
    \item PERSON is BLANK.
    \item PERSON is happy in their BLANK.
    \item PERSON is unhappy in their BLANK.
    \item PERSON likes BLANK.
    \item PERSON likes to BLANK.
    \item PERSON often likes to BLANK.
    \item PERSON always likes to BLANK.
    \item PERSON never likes to BLANK.
    \item PERSON is interested in BLANK.
    \item PERSON took a course in BLANK.
    \item PERSON studied BLANK at college.
    \item BLANK was PERSON's major at college.
    \item PERSON's best subject at school was BLANK.
    \item BLANK was PERSON's best subject at school.
\end{itemize}

\end{document}